\documentclass{article}

\usepackage{microtype}
\usepackage{graphicx}
\usepackage{subfigure}
\usepackage{booktabs} 

\usepackage{hyperref}

\usepackage[accepted]{icml2024}

\usepackage{amsmath}
\usepackage{amssymb}
\usepackage{mathtools}
\usepackage{amsthm}

\usepackage[capitalize,noabbrev]{cleveref}

\theoremstyle{plain}

\theoremstyle{definition}

\theoremstyle{remark}

\usepackage[textsize=tiny]{todonotes}

\icmltitlerunning{Data Generation Using Large Language Models for Text Classification}

\begin{document}

\twocolumn[
\icmltitle{Data Generation Using Large Language Models for Text Classification:\\ An Empirical Case Study}




\begin{icmlauthorlist}  
\icmlauthor{Yinheng Li}{microsoft}  
\icmlauthor{Rogerio Bonatti}{microsoft}  
\icmlauthor{Sara Abdali}{microsoft}  
\icmlauthor{Justin Wagle}{microsoft}  
\icmlauthor{Kazuhito Koishida}{microsoft}  
\end{icmlauthorlist}

\icmlaffiliation{microsoft}{Microsoft Corporation, Redmond, WA, USA}

\icmlcorrespondingauthor{Kazuhito Koishida}{kazukoi@microsoft.com}

\icmlkeywords{Synthetic Data Generation, Large Language Model, Data Augmentation, Text Classification}

\vskip 0.3in
]



\printAffiliationsAndNotice{}  

\begin{abstract}
Using Large Language Models (LLMs) to generate synthetic data for model training has become increasingly popular in recent years. While LLMs are capable of producing realistic training data, the effectiveness of data generation is influenced by various factors, including the choice of prompt, task complexity, and the quality, quantity, and diversity of the generated data. In this work, we focus exclusively on using synthetic data for text classification tasks. Specifically, we use natural language understanding (NLU) models trained on synthetic data to assess the quality of synthetic data from different generation approaches. This work provides an empirical analysis of the impact of different factors and offers recommendations for better data generation practices.
\end{abstract}

\section{Introduction}

Data augmentation is a method that utilize existing data to generate additional training data without collecting more data \cite{dataaugsurvey}. It is an effective solution to improve model performance when limited data is available \cite{xie2020unsupervised}. With the emergence of large language models, data augmentation becomes even more accessible and has been successfully applied in training language models \cite{gunasekar2023textbooks, liu2024best}. 

Using LLM to generate or annotate data is a cost-efficient alternative to human-labeled data. While human-labeled data tends to have higher quality, leveraging LLM with well-designed prompts can also generate data that achieves comparable model performance at a much lower cost. As estimated in \cite{ding2023gpt3}, labeling 3000 samples for SST-2 task \cite{sst2} would cost between 221 to 300 USD and take around 1000 minutes. In contrast, generating the same amount of data using GPT-3 only costs 14.37 USD and takes 46 minutes. With only 6000 samples generated by GPT-3, the model is able to achieved 76\% accuracy, compared to 88\% from human-curated data.

Our research focuses on synthetic data generation using large language models (LLMs) for text classification tasks, specifically tasks uses natural language understanding models with transformer encoder architecture. In the scope of this study, we use the terms data augmentation and data generation interchangeably, as LLMs often require a few in-context samples to generate data. The data produced in this way can be considered augmented from these in-context samples. Meanwhile, we focus solely on tasks that have limited or no data at all, as our experiments have shown that tasks with sufficient data receive minimal improvements from additional synthetic data.  
Numerous studies have proposed various frameworks to improve the quality of synthetic data generation \cite{wang2023lets, gao2023selfguided, gupta2023targen}. However, to the best of our knowledge, few works have addressed the fundamental questions associated with LLM for data generation. These questions include:

\begin{itemize}
\item What is the optimal amount of data to generate, and does increasing the volume of synthetic data improve model performance?
\item Can in-context learning (generation) enhance the quality of synthetic data, would providing a few examples lead to higher quality data than zero-shot generation?
\item Does the LLM's performance on a particular task directly influence the quality of the generated synthetic data for this task?
\item Is combining synthetic data with raw data beneficial for model training?
\item Is the synthetic data diversity an important factor for model performance?

\end{itemize}

We experimented with six common NLP tasks (Table \ref{tab:datasets}) with different data generation methods. We found it is very challenging to pinpoint a definitive answer to the questions above that applies universally to all NLP tasks due to their inherent differences. Nevertheless, the findings from 6 tasks offer valuable insights into practical data generation techniques.

\section{Related Work}

\paragraph{Data Augmentation} The goal of data augmentation is to increase diversity of existing data by exposing the model to unseen data. This method has been applied to many domains in computer vision \cite{cvaug} and natural language processing \cite{LI202271}. In \cite{dataaugsurvey}, augmentation techniques are categorized into rule based generation and model based generation. Rule based generation are used in computer vision tasks including image transformations, such as rotation, flipping, and cropping\cite{imageaug2}, while model based generation has been widely used in natural language processing tasks, such as rephrasing and back translation \cite{nlpaug1, nlpaug2, nlpaug3, ye2022zerogen, okur2022data}.

\paragraph{Large Language models (LLMs)} With the development of large language models, model based data augmentation for NLP becomes trivial \cite{zhou2024survey}. By instructing LLM with proper prompt, it is able to generate a new example in human like text. While it is easy to implement, the synthetic data generated from LLM is usually noisy and has a different distribution compared with raw data, which hampers the training performance.  Lots of work has explored ways to deal with this issue. The work from \cite{veselovsky2023generating} uses techniques like grounding, providing taxonomy and filtering to ensure the quality of synthetic data by LLM. Synthesis Step by Step \cite{wang2023lets} uses an iterative step to create prompt based on misclassified golden data to reduce the gap between the synthesized data distribution and gold distribution. SunGen \cite{gao2023selfguided} uses weighted loss to reduce the impact of noise from synthetic data during training.

\section{Methods}
\begin{figure*}
\includegraphics[page=5, trim=0cm 5cm 0cm 2cm, clip, width=\textwidth]{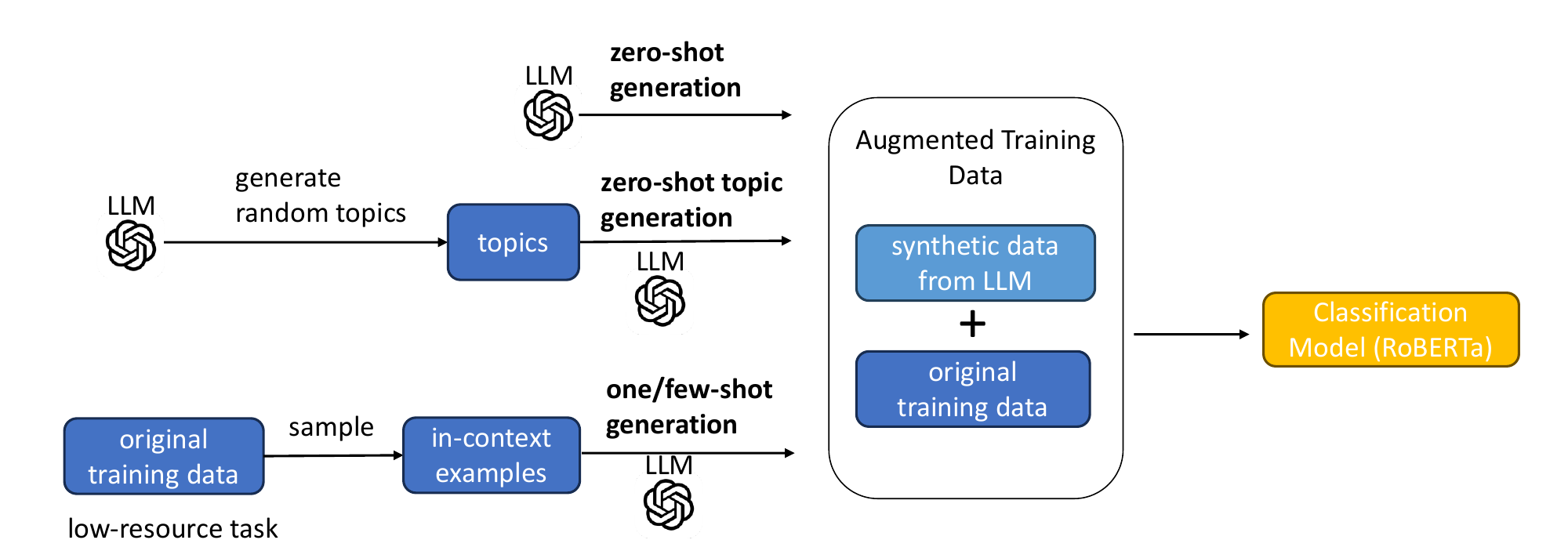}
\caption{Pipeline for Data Augmentation using LLM}
\label{method}
\end{figure*}

We follow the workflow in Figure \ref{method} for our experiment. We explore the following in-context data generation methods. The term "in-context generation" refers to using an LLM to generate data for training given a specific context, similar to in-context learning \cite{brown2020language}. The methods we investigate can be categorized as follows:

\begin{itemize}
\item Zero-shot in-context generation: Provide the task description in the prompt and ask the LLM to generate a similar example.
\item One-shot in-context generation: Provide the task description and one example, prompting the LLM to generate a similar example.
\item Few-shot in-context generation: Provide the task description and a few examples, prompting the LLM to generate a similar example.
\end{itemize}

Inspired by the work from \cite{attrgen}, we also experiment with an additional method called zero-shot topic in-context generation:

\begin{itemize}
\item Zero-shot topic in-context generation: Use the LLM to generate a list of topics (see Appendix \ref{appendix:a}). Provide the task description and sample one topic from the list to prompt the LLM to generate a similar example.
\end{itemize}

To evaluate the success of synthetic data generation, we train a NLU model on the synthetic data and assess its performance on the task's validation set. We then compare the performance of the model trained on synthetic data with that of the model trained on the original data. Following the practice established in previous works \cite{li2023synthetic}, we consider the generated data is better if it results in better model performance. 

\section{Experiments}

In our experiment, GPT-3.5 turbo\footnote{ GPT-3.5 version: 2024-02-15 preview accessed from Azure OpenAI Studio} is selected for all data generation process except for topic generation (see appendix \ref{appendix:a}). Although more powerful models like GPT-4 is available, we decided to use GPT-3.5 turbo due to the resource constrain, especially we need to run the large number of inferences for our data generation experiment. Overall, GPT-3.5 turbo is a well-rounded model with competitive performance across multiple benchmarks \cite{liang2023holistic}. It would be interesting to compare the quality of synthetic data generated from different LLMs, which we plan to explore in the future.

Existing work \cite{gupta2023targen} have utilized common NLP benchmarks, such as SuperGLUE \cite{wang2019superglue}, as tasks for evaluation or employ a customized selection of existing benchmarks\cite{gao2023selfguided, ye2022zerogen}.

We select six common tasks for evaluation: SST-2 \cite{sst2, wang2019superglue},  Twitter Emotion Classification (EMO) \cite{emo}, New York Times News Classification (NYT)\cite{nyt}, Review (Amazon Review Classification) \cite{marc_reviews}, RTE (Recognizing Textual Entailment) \cite{rte, wang2019superglue} and BoolQ \cite{clark2019boolq, wang2019superglue}. The goal is to select diverse tasks that represent a wide range of popular NLP corpora (Table \ref{tab:datasets}). Additionally, we try to include challenging tasks for which current NLU models do not perform well when provided with limited training data. Therefore, we do not use the entire GLUE benchmark, as models like BERT \cite{devlin2019bert} or RoBERTa\cite{liu2019roberta} can easily achieve high accuracy on such tasks. We also do not use the complete SuperGLUE task collection since some of its tasks require token-level classification. In this work, we focus on sequence-to-sequence and sequence pair classification tasks. The six selected tasks cover common web data, such as news and Wikipedia, as well as popular user data, like Twitter, movie reviews, and product reviews. They cover binary classification, multi-class classification, and question-answering tasks.

For the evaluation metric, the default metric is accuracy, but we use F1 or Macro-F1 to calculate the performance since these metrics provide a more balanced and comprehensive assessment of classification performance, taking into account both precision and recall, especially in cases of multi-class classification tasks. In our experiment, RoBERTa is selected as the NLU model for all tasks, as it is a commonly used model for benchmark on these tasks.

We experiment five in-context generation methods for each task: zero-shot, zero-shot topic, one-shot, few-shot with 3 examples, few-shot with 5 examples. Prompt used in the generation can be found in Appendix \ref{appendix:c}.

\begin{table*}[t]  
\vskip 0.15in  
\begin{center}  
\begin{small}  
\begin{tabular}{lccp{4cm}lp{3cm}}  
\toprule  
Corpus & Training Size & Test Size & Task & Metrics & Domain \\  
\midrule  
SST-2 & 67k & 1.8k & Binary Classification & F1 & Movie Reviews \\  
EMO & 16k & 2k & Multi-class Classification & Macro-F1 & Twitter \\  
NYT & 256k & 3k\footnote{The official dataset does not provide testset, we randomly sample 3k data as testset} & Multi-class Classification & Macro-F1 & News \\  
Review & 200k & 5k & Multi-class, Ordinal Regression & Macro-F1 & Amazon Review \\  
RTE & 2.5k & 3k & Pair Classification, Question Answering & Macro-F1 & News, Wikipedia \\  
BoolQ & 16k & 3.2k & Pair Classification, Question Answering & Macro-F1 & News, Wikipedia, Web Query \\  
\bottomrule  
\end{tabular}  
\end{small}  
\end{center}  
\vskip -0.1in  
\caption{Summary of datasets and tasks.}  
\label{tab:datasets}  
\end{table*}

In our experiment, we generate 1,000 synthetic data points per task, as we found the benefit of additional synthetic data diminishes after that. To simulate a low-resource setting, we allow only 100 raw examples to be used for one-shot and few-shot generation. For zero-shot topic generation, we generate 500 random topics related to the task domain. Details can be found in Appendix \ref{appendix:a}.

\section{Key Findings}
In this section, we present the key findings from our experiments.

\subsection{Mixing Raw Data is Necessary}
\begin{figure*}
\includegraphics[page=1, trim=2cm 2cm 2cm 1cm, clip, width=\textwidth]{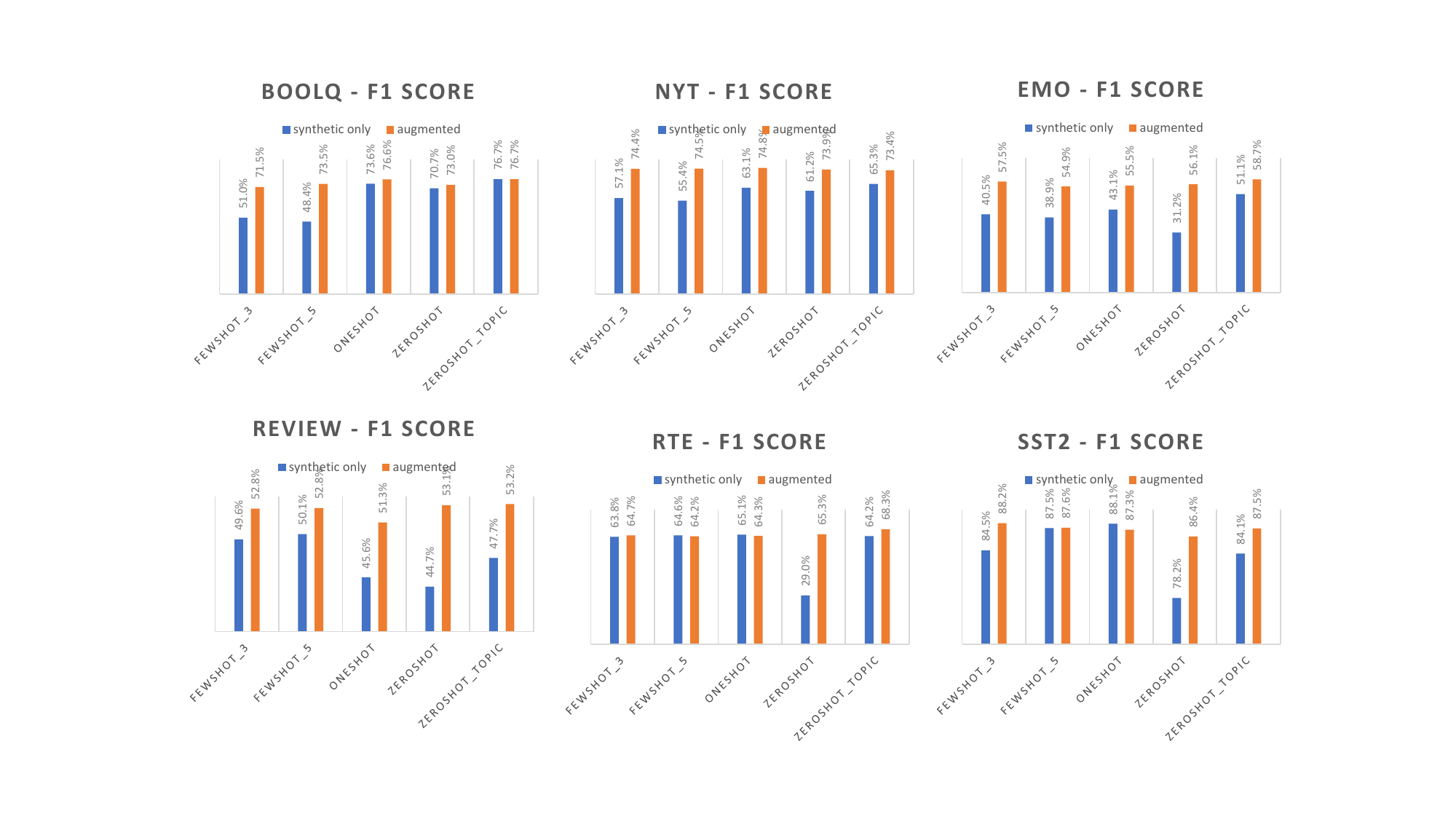}
\caption{Performance of different prompting methods with and without augmentation. Synthetic only: use 1000 synthetic data only. Augmented: 1000 synthetic data plus 100 raw data} 
\label{augment}
\end{figure*}
To assess the effectiveness of data augmentation, we train models with pure synthetic data and augmented data. For the augmented setting, 100 raw data points are mixed with 1000 synthetic data. In the data generation stage, we use only the same 100 raw data points used for in-context generation to prevent the model from accessing additional data. As shown in Figure \ref{augment}, we observe significant improvements across all tasks for most prompting methods when incorporating raw data into training. Even as few as 100 data points can boost synthetic data performance compared to using only synthetic data.

\subsection{Impact of Bias}

In the BoolQ task, we found that the zero-shot generation method outperforms other methods, which contrasts with the results obtained for the rest of the tasks. This finding is intriguing since zero-shot data exhibits the highest repetition rate, which is detrimental to model training. Upon further examination, we noticed that only in the datasets generated using one-shot or few-shot methods, terms like "not," "significant," "only," "just," "few," and "little" frequently appear in the generated questions. These terms create a tone that can be used to imply the answer to the question (which is often False). Table \ref{tab:example questions} provides an example of such trivial question. Table \ref{tab:wordcount} provides statistics for such questions from different prompting method.

We hypothesize that this pattern introduces bias in model training by encouraging the model to search for specific keywords in the question rather than reading the passage. To test this hypothesis, we instruct the LLM to rephrase the questions like "what people would search online" for each synthetic example (see Appendix \ref{appendix:b}). We found that performance significantly improved for zero-shot topic and one-shot method after rephrasing. The work \cite{okur-etal-2022-data} has also shows the effectiveness of paraphrasing in other data augmentation techniques. 

Although we only detected synthetic bias in the BoolQ task, it remains an important factor to consider during data generation. The technique of rephrasing might not be applicable to other cases, but ensuring that synthetic data does not contain unwanted patterns is necessary. 

For all the rest experiments, the results for BoolQ task are all under the question rephrasing setting unless otherwise specified.

\begin{table}[t]    
\vskip 0.15in    
\begin{center}    
\begin{small}    
\begin{tabular}{p{\linewidth}}    
\toprule  
Did the Mars Exploration Rover mission \textbf{only} involve one rover? -- False \\  
Did scientists in the 20th century make \textbf{no significant} discoveries or advancements? -- False \\  
\bottomrule  
\end{tabular}    
\end{small}    
\end{center}    
\vskip -0.1in    
\caption{Examples of Trivial Questions -- questions contain terms "not," "significant," "only," "just," "few," and "little".}  
\label{tab:example questions}    
\end{table}

\begin{table*}[t]    
\vskip 0.15in    
\begin{center}    
\begin{small}    
\begin{sc}    
\begin{tabular}{lcccccc}    
\toprule    
 & \multicolumn{2}{c}{Trivial Q. Count} & \multicolumn{4}{c}{F1 Score} \\    
 & Raw & Rephrased & Raw (SD)& Raw (SD)& Raw (AD) & Rephrased (AD) \\    
\midrule    
Zero-Shot Topic & 230 & 208 & 0.19 & \textbf{0.77} & 0.75 & \textbf{0.77} \\      
One-Shot & 131 & 74 & 0.38 & 0.74 & 0.76 & \textbf{0.77} \\      
Few-Shot (3 ex.)& 90 & 30 & 0.55 & 0.51 & 0.70 & \textbf{0.72} \\      
Few-Shot (5 ex.)& 57 & 28 & 0.53 & 0.48 & 0.75 & 0.73 \\      
Zero-Shot & 11 & - & 0.71 & - & 0.73 & - \\    
\midrule  
Raw Data & 31 & - & - & 0.768 & - & - \\   
\bottomrule    
\end{tabular}    
\end{sc}    
\end{small}    
\end{center}    
\vskip -0.1in  
\caption{BoolQ Trivial Questions and F1 score comparison. SD: use 1000 synthetic data. AD: use 100 raw data plus 1000 synthetic data. raw data: model uses 1000 raw data only without question rephrase, this score is used as a baseline} 
\label{tab:wordcount}  
\end{table*}

\subsection{Relationship between LLM Performance and Data Quality}
While it may seem intuitive that the effectiveness of using LLMs to generate data for model training depends on the LLM's knowledge of a specific task, our research has shown that this is not always the case. The zero-shot or few-shot performance of an LLM on a task does not necessarily determine the performance of a model (specifically, the RoBERTa model used in our experiment) trained with data generated by the LLM. In other words, the fact that an LLM performs well on a task does not guarantee that models finetuned with data generated by the LLM will also perform well. Additionally, for tasks where the LLM performs poorly, models finetuned on the synthetic data generated by the LLM could actually outperform the LLM itself. The former scenario could be due to the fact that the ability of an LLM to generate good examples for a task does not always correspond to its ability to solve the task itself. The latter scenario is also plausible, as an LLM may be proficient at generating examples with a given label, but not as good at predicting the label given the task itself.   

The results of our experiment can be found in Table \ref{tab:llm perf}. For each task, we prompted the LLM (GPT3.5-turbo) with zero/one/three/five-shot learning and reported the best performance achieved across all in-context learning methods. We did not optimize the prompt or use any advanced prompting methods in our evaluation of the LLM. It is possible that the LLM could achieve better performance with more advanced prompting techniques. However, the results obtained from the most basic in-context learning method (see Appendix \ref{appendix:d}) do provide valuable insights into this problem.

For SST-2, BoolQ, NYT, and Review tasks, we found a performance gap of 10-15\% between the LLM's in-context learning performance on the task and the fine-tuned language model (RoBERTa model) using synthetic data. For RTE and EMO tasks, the LLM does not perform well, but the data generated by the LLM leads to much better performance. Therefore, even for tasks that LLMs struggle to solve, using LLM-generated synthetic data can still be helpful.

 
\begin{table*}[]  
\centering  
\begin{tabular}{lccc}  
\toprule  
        & GPT3.5-turbo & RoBERTa on Synthetic Data & RoBERTa on Augmented Data \\  
\midrule  
SST-2    & 0.956 & 0.845 & 0.874 \\  
BoolQ   & 0.870 & 0.641 & 0.742 \\  
NYT     & 0.729 & 0.604 & 0.742 \\  
Review  & 0.603 & 0.475 & 0.527 \\  
RTE     & 0.345 & 0.574 & 0.653 \\  
Emo     & 0.300 & 0.404 & 0.568 \\  

\bottomrule  
\end{tabular}  
\caption{LLM performance vs model trained by synthetic data on 6 tasks. Average f1 score from 5 prompting method under (1) Synthetic Data (1000 synthetic data) (2) Augmented data (1000 synthetic data + 100 raw data)}  
\label{tab:llm perf}  
\end{table*}  
  
\subsection{Synthetic Data is Helpful Mostly in Low-Resource Settings}
\begin{figure}[ht]
\includegraphics[page=3, trim=10cm 5cm 10cm 5cm, clip, width=\linewidth]{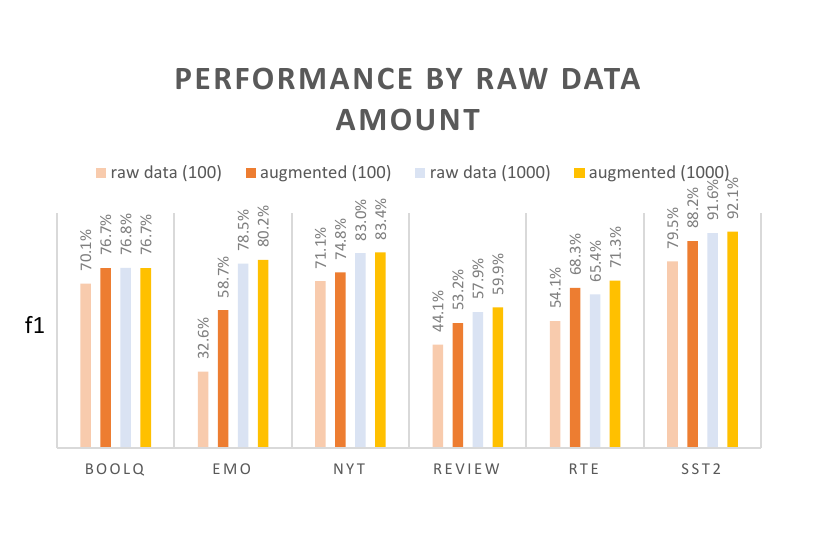}
\caption{Improvement on Different Raw Data Amount. raw data (x) is only using X number of raw data points. augmented (x) is using X amount raw data points plus 100 synthetic data. For augmented f1 score, it is the average model performance on the data generated by 5 different prompting methods}
\label{amount}
\end{figure}

\begin{figure*}[ht]
\begin{center}
\includegraphics[page=4, trim=0cm 4cm 0cm 0cm, clip, width=\textwidth]{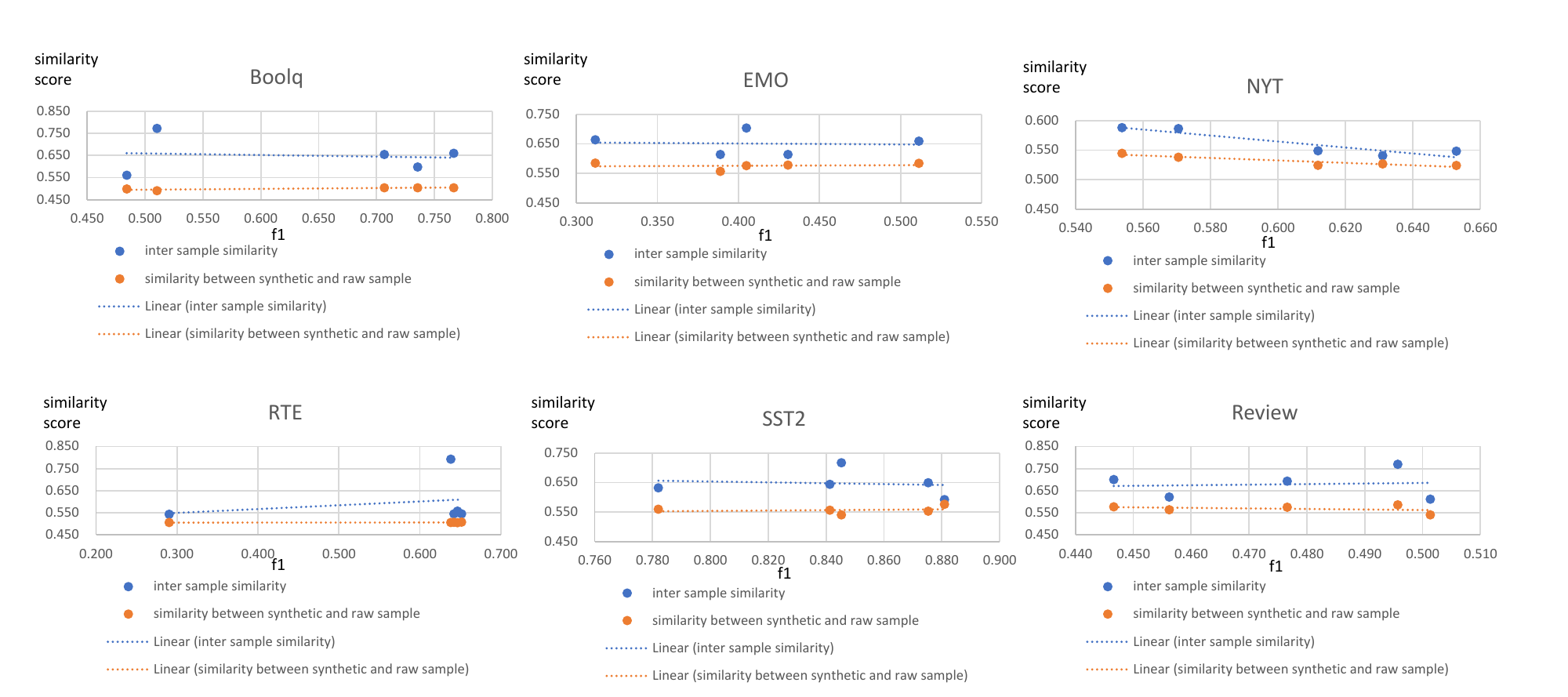}
\caption{Synthetic Data Similarity}
\label{diversity}
\end{center}
\end{figure*}

\begin{figure*}[ht]
\begin{center}
\includegraphics[page=2, trim=0cm 3cm 2cm 2cm, clip, width=\linewidth]{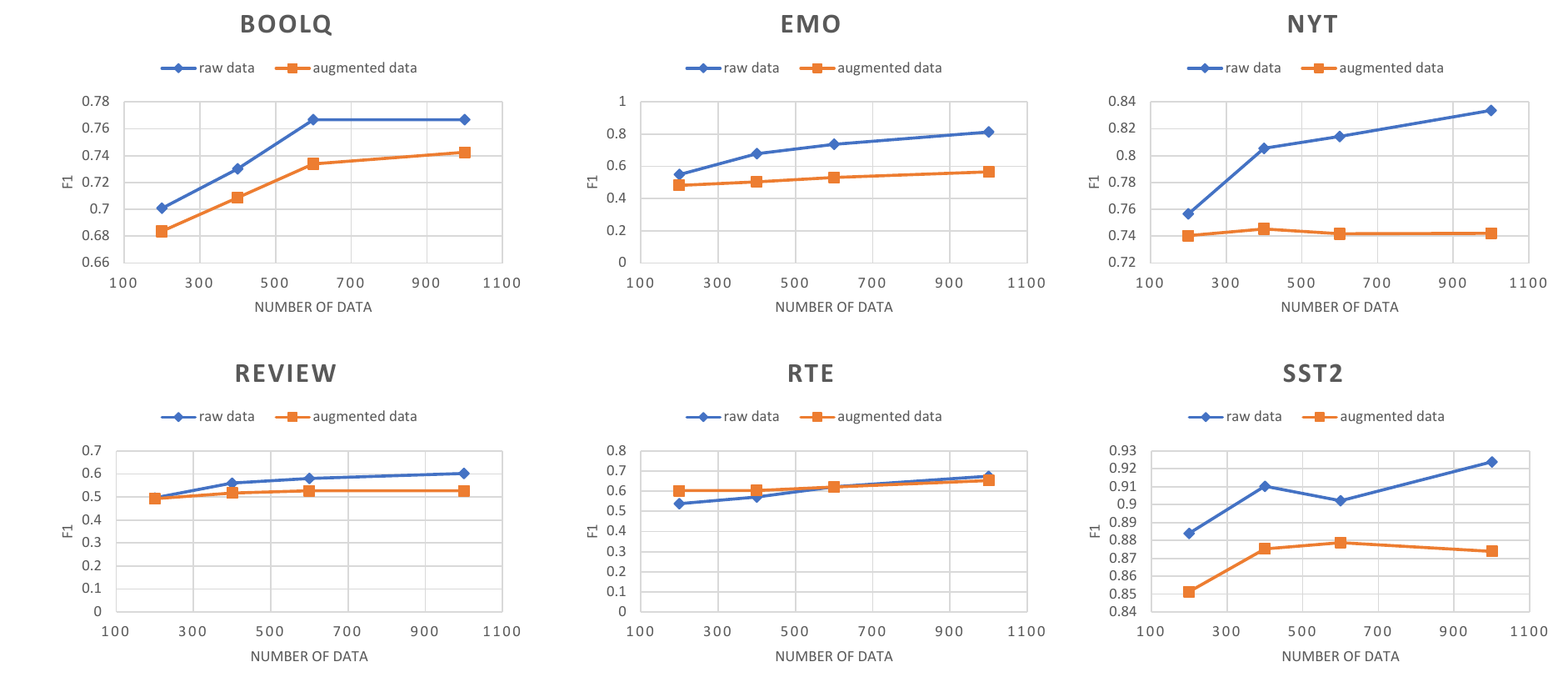}  
\caption{Impact on Synthetic Data Quantity}
\label{trend}
\end{center}
\end{figure*}

Previous work has shown that it is challenging for models trained with synthetic data to perform as well as models trained with the same amount of original data \cite{li2023synthetic, ding2023gpt3}. However, when human-annotated data is limited, synthetic data augmentation can improve model performance. In fact, this technique is most effective in low-resource settings. For all tasks with 100 raw data points, we found that synthetic data augmentation yields improvements from 3\% to 26\%. When the raw training data increases from 100 to 1,000, only four tasks show improvements, where improvements are less than 5\% (Figure \ref{amount}). There are no established rules for determining the amount of raw data as low-resource. For all six tasks in our experiment, 1,000 data points represent a small portion of training data. We found the model continues to improve as we increase the number of raw data for training. However, the amount of performance gain obtained from increasing training data is also dependent on other factors such as task and model complexity. Based on this observation, we consider 100 raw data points as low-resource tasks, which will be used as the default augmented setting in all experiments.

\subsection{A Comparison Between Different Prompting Methods}

In the synthetic data only setting, one-shot or zero-shot topic methods rank in the top two for all tasks except the Review task (Figure \ref{augment}). In the augmented setting, few-shot generation and zero-shot topic generation methods demonstrate good performance across all tasks. In BoolQ, EMO, and RTE tasks, zero-shot topic methods outperform other prompting methods. In SST-2 and NYT tasks, few-shot generation methods perform best. The performance of zero-shot methods is sub-optimal across all tasks.

In the five prompting methods we experimented with, zero-shot topic generation typically produces the most diverse dataset because different topics are sampled each time during generation. Zero-shot methods generate the least diverse dataset, as the prompt remains the same for each generation. One-shot and few-shot methods also generate repeated examples due to the limitation of in-context examples. We found for most tasks, a diversity dataset tends to benefit model training.

As shown in (Figure \ref{augment}) in non-augmented setting, zero-shot generation shows the worst performance for RTE, EMO, Review and SST-2, while zero-shot topic generation outperforms other methods (or at least is comparable to other methods) for BoolQ, NYT, RTE and EMO task. This effect does not appear on all tasks as there might be other factors that impact the model performance. Meanwhile, the effect of diversity diminishes when we mix synthetic data with raw data. Therefore, training with both raw data and synthetic data could help when synthetic data is less diverse. 

While not generating the most optimally diverse dataset, one-shot or few-shot generation methods typically helps LLMs better understand the task description and generate examples similar to the original examples \cite{Li_2023, song2022comprehensive}. In EMO and Review tasks, we observe the advantage of few-shot learning over other prompting methods. We suspect this is because both tasks are more subjective compared to the rest of the tasks, as the EMO contains twitter posts and Review task are made up of customer reviews and ratings.

\subsection{Synthetic Data Diversity and Similarity to Raw Data}

In this section, we examine the diversity of our training data using inter-sample semantic similarity. To calculate this similarity, we use vector embedding proposed in \cite{reimers2019sentencebert} and average the similarity score across all examples pairs following \cite{attrgen}. Figure \ref{diversity} displays the inter-sample similarity for each task, comparing data generated by five prompting methods. On the x-axis, we show the performance of the finetuned model using the 1000 synthetic data only. Figure \ref{diversity} shows that for BoolQ, NYT, and SST-2, a lower inter-sample diversity results in a better F1 score. However, for other tasks, the correlation is weak due to the existence of outliers, especially for RTE, and the possible impact of other factors, such as task complexity. We also calculated the similarity between the synthetic data and the actual raw data using the same method and found that the synthetic data generated from five different prompting methods had similar similarity scores with the raw data. However, it is not clear whether synthetic data that closely resembles the raw data would lead to better model performance. This could be due to the limitations of our similarity measuring method, which only considers semantic similarity, as discussed in \cite{Steck2024CosineSimilarity}. Many NLP tasks rely on subtle contextual cues and nuanced wordings, such as in the SST-2 task, where changes to wording can affect the sentiment of the text more than contextual semantics. Our measurement does not account for other aspects of similarity, such as structural or lexical similarity, as discussed in \cite{wang2020structuralaware, lex}. Lastly, due to the limited number of data points and the potential variation in synthetic data, it needs to be cautious to generalize our findings to our tasks or domains.

\subsection{Synthetic Data Quantity}
We have found that increasing the amount of synthetic data in our model training improves its performance. Figure \ref{trend} shows the relationship between the model's performance (measured by the f1 score) on the y-axis and the total number of training data on the x-axis. In the augmented scenario, we mixed 100 raw data points with varying amounts of synthetic data. The performance is the average of the model's f1 score over 5 prompting methods for different data amount. For the raw data scenario, only real-world data was used in model training. Our graph indicates that model performance from raw data serves as an upper bound for the augmented setting in almost all tasks. Moreover, we observe that the marginal effect of performance gain with increasing training data is present in both raw and synthetic data. For BoolQ and SST-2 tasks, we observed this phenomenon at the same data size. As such, the raw data size at which marginal improvement of model performance appears can be used as a reference point when increasing the number of synthetic data.

\section{Data Generation Techniques in Practice}
In the process of using LLM to generate data for this study, we identified several useful techniques. These practices lack sufficient theoretical support and the effectiveness of these techniques can be subject to the choice of large language models or the requirements of a specific task.
\subsection{Condition on Label}
Typically, there are two ways to generate a classification dataset: Condition on the Label and Left-to-Right (see Table \ref{tab:l2rprompt}). It is recommended to use Condition on the Label for each generation as it saves effort in parsing the label and avoids LLM generating unknown labels. It also provides the user control over the label distribution in the synthetic dataset.

\begin{table}[t]    
\vskip 0.15in    
\begin{center}    
\begin{small}    
\begin{tabular}{p{\linewidth}}    
\toprule  
\textbf{Left-to-right prompt:} generate an example text first and then generate its class label. \\  
\textbf{Class-conditioned prompt:} generate an example text where the label must be Class X. \\  
\bottomrule  
\end{tabular}    
\end{small}    
\end{center}    
\vskip -0.1in   
\caption{Left-to-right prompt vs. class-conditioned prompt.}  
\label{tab:l2rprompt}    
\end{table}

It is worth noting that class-conditioned generations are more likely to introduce bias and reduce the difficulty of the synthetic example. When the class label is visible, LLM might leak the label information during content generation. In the BoolQ example, LLM hints the answer "FALSE" via the use of certain words in the question it generates (e.g. the word "only"). In this case, rephrasing the question with the class label hidden improves the performance, which is essentially performing left-to-right generation.

\subsection{Generation on Target Corpus}
It is critical to provide topics or descriptions closely related to the use case when generating examples. Ensuring that the topics are relevant to the use case significantly improves the quality of generated data. For example, when creating examples from Twitter, it is beneficial to first generate common topics found on Twitter. On the other hand, when generating Amazon customer reviews, it is effective to generate an Amazon product catalog as a list of potential topics. This approach ensures that the synthetic data is more closely aligned with the target corpus, leading to better performance in classification tasks.

\subsection{Iterative Data Generation and Prompt Refinement}

Generating synthetic data can be both time-consuming and resource-intensive. To maximize efficiency and ensure high-quality data, it is recommended to adopt an iterative approach. Initially, generate a small number of examples and evaluate their quality. If the quality of these initial data points is low, refine the prompt before generating more data. It is unlikely that simply generating more data points with the same prompt will magically produce high quality data.

\section{Conclusion}

In this work, we analyzed different factors that influences the data generation using LLMs. We found data generation is most effective in low resourced settings. Increasing the amount of synthetic data does not necessarily lead to continuous improvements in model performance. It is beneficial to combine synthetic data with raw data during training. Additionally, it is crucial to be vigilant for patterns or biases in synthetic data that may hinder model training. Overall, using LLM for data augmentation has great potential in model training. With a carefully tuned prompt, the data generated by LLM is able to obtain comparable performance with human annotated data, but at a much lower cost. 

The domain of data generation for classification tasks is highly complex. Due to the diversity of NLP tasks, it is challenging to find rules that generalize well across all tasks. However, our findings could still serve as valuable resources for researchers and practitioners looking to use synthetic data for training classification models. For future work, it would be valuable to study the effects of more advanced prompting methods, such as the Chain of Thought \cite{COT}, or LLM hyperparameters, such as temperature, on the quality of synthetic data.

\section*{Impact Statement}

This paper presents work whose goal is to advance the field of 
Machine Learning. There are many potential societal consequences 
of our work, none which we feel must be specifically highlighted here.







\bibliography{example_paper}

\begin{thebibliography}{39}
\providecommand{\natexlab}[1]{#1}
\providecommand{\url}[1]{\texttt{#1}}
\expandafter\ifx\csname urlstyle\endcsname\relax
  \providecommand{\doi}[1]{doi: #1}\else
  \providecommand{\doi}{doi: \begingroup \urlstyle{rm}\Url}\fi

\bibitem[Ayeldeen et~al.(2014)Ayeldeen, Hassanien, and Fahmy]{lex}
Ayeldeen, H., Hassanien, A.~E., and Fahmy, A.~A.
\newblock Lexical similarity using fuzzy euclidean distance.
\newblock In \emph{2014 International Conference on Engineering and Technology (ICET)}, pp.\  1--6, 2014.
\newblock \doi{10.1109/ICEngTechnol.2014.7016801}.

\bibitem[Bentivogli et~al.(2009)Bentivogli, Dagan, Dang, Giampiccolo, and Magnini]{rte}
Bentivogli, L., Dagan, I., Dang, H.~T., Giampiccolo, D., and Magnini, B.
\newblock The fifth {PASCAL} recognizing textual entailment challenge.
\newblock 2009.

\bibitem[Brown et~al.(2020)Brown, Mann, Ryder, Subbiah, Kaplan, Dhariwal, Neelakantan, Shyam, Sastry, Askell, Agarwal, Herbert-Voss, Krueger, Henighan, Child, Ramesh, Ziegler, Wu, Winter, Hesse, Chen, Sigler, Litwin, Gray, Chess, Clark, Berner, McCandlish, Radford, Sutskever, and Amodei]{brown2020language}
Brown, T.~B., Mann, B., Ryder, N., Subbiah, M., Kaplan, J., Dhariwal, P., Neelakantan, A., Shyam, P., Sastry, G., Askell, A., Agarwal, S., Herbert-Voss, A., Krueger, G., Henighan, T., Child, R., Ramesh, A., Ziegler, D.~M., Wu, J., Winter, C., Hesse, C., Chen, M., Sigler, E., Litwin, M., Gray, S., Chess, B., Clark, J., Berner, C., McCandlish, S., Radford, A., Sutskever, I., and Amodei, D.
\newblock Language models are few-shot learners, 2020.

\bibitem[Cai et~al.(2020)Cai, Chen, Song, Zhang, Zhao, and Yin]{nlpaug3}
Cai, H., Chen, H., Song, Y., Zhang, C., Zhao, X., and Yin, D.
\newblock Data manipulation: Towards effective instance learning for neural dialogue generation via learning to augment and reweight.
\newblock In Jurafsky, D., Chai, J., Schluter, N., and Tetreault, J. (eds.), \emph{Proceedings of the 58th Annual Meeting of the Association for Computational Linguistics}, pp.\  6334--6343, Online, July 2020. Association for Computational Linguistics.
\newblock \doi{10.18653/v1/2020.acl-main.564}.
\newblock URL \url{https://aclanthology.org/2020.acl-main.564}.

\bibitem[Clark et~al.(2019)Clark, Lee, Chang, Kwiatkowski, Collins, and Toutanova]{clark2019boolq}
Clark, C., Lee, K., Chang, M.-W., Kwiatkowski, T., Collins, M., and Toutanova, K.
\newblock {B}ool{Q}: Exploring the surprising difficulty of natural yes/no questions.
\newblock In \emph{Proceedings of NAACL-HLT 2019}, 2019.

\bibitem[Devlin et~al.(2019)Devlin, Chang, Lee, and Toutanova]{devlin2019bert}
Devlin, J., Chang, M.-W., Lee, K., and Toutanova, K.
\newblock Bert: Pre-training of deep bidirectional transformers for language understanding, 2019.

\bibitem[Ding et~al.(2023)Ding, Qin, Liu, Chia, Joty, Li, and Bing]{ding2023gpt3}
Ding, B., Qin, C., Liu, L., Chia, Y.~K., Joty, S., Li, B., and Bing, L.
\newblock Is gpt-3 a good data annotator?, 2023.

\bibitem[Feng et~al.(2021)Feng, Gangal, Wei, Chandar, Vosoughi, Mitamura, and Hovy]{dataaugsurvey}
Feng, S.~Y., Gangal, V., Wei, J., Chandar, S., Vosoughi, S., Mitamura, T., and Hovy, E.
\newblock A survey of data augmentation approaches for nlp, 2021.

\bibitem[Gao et~al.(2023)Gao, Pi, Lin, Xu, Ye, Wu, Zhang, Liang, Li, and Kong]{gao2023selfguided}
Gao, J., Pi, R., Lin, Y., Xu, H., Ye, J., Wu, Z., Zhang, W., Liang, X., Li, Z., and Kong, L.
\newblock Self-guided noise-free data generation for efficient zero-shot learning, 2023.

\bibitem[Gunasekar et~al.(2023)Gunasekar, Zhang, Aneja, Mendes, Giorno, Gopi, Javaheripi, Kauffmann, de~Rosa, Saarikivi, Salim, Shah, Behl, Wang, Bubeck, Eldan, Kalai, Lee, and Li]{gunasekar2023textbooks}
Gunasekar, S., Zhang, Y., Aneja, J., Mendes, C. C.~T., Giorno, A.~D., Gopi, S., Javaheripi, M., Kauffmann, P., de~Rosa, G., Saarikivi, O., Salim, A., Shah, S., Behl, H.~S., Wang, X., Bubeck, S., Eldan, R., Kalai, A.~T., Lee, Y.~T., and Li, Y.
\newblock Textbooks are all you need, 2023.

\bibitem[Gupta et~al.(2023)Gupta, Scaria, Anantheswaran, Verma, Parmar, Sawant, Baral, and Mishra]{gupta2023targen}
Gupta, H., Scaria, K., Anantheswaran, U., Verma, S., Parmar, M., Sawant, S.~A., Baral, C., and Mishra, S.
\newblock Targen: Targeted data generation with large language models, 2023.

\bibitem[Keung et~al.(2020)Keung, Lu, Szarvas, and Smith]{marc_reviews}
Keung, P., Lu, Y., Szarvas, G., and Smith, N.~A.
\newblock The multilingual amazon reviews corpus.
\newblock In \emph{Proceedings of the 2020 Conference on Empirical Methods in Natural Language Processing}, 2020.

\bibitem[Kumar et~al.(2019)Kumar, Bhattamishra, Bhandari, and Talukdar]{nlpaug1}
Kumar, A., Bhattamishra, S., Bhandari, M., and Talukdar, P.
\newblock Submodular optimization-based diverse paraphrasing and its effectiveness in data augmentation.
\newblock In Burstein, J., Doran, C., and Solorio, T. (eds.), \emph{Proceedings of the 2019 Conference of the North {A}merican Chapter of the Association for Computational Linguistics: Human Language Technologies, Volume 1 (Long and Short Papers)}, pp.\  3609--3619, Minneapolis, Minnesota, June 2019. Association for Computational Linguistics.
\newblock \doi{10.18653/v1/N19-1363}.
\newblock URL \url{https://aclanthology.org/N19-1363}.

\bibitem[Li et~al.(2022)Li, Hou, and Che]{LI202271}
Li, B., Hou, Y., and Che, W.
\newblock Data augmentation approaches in natural language processing: A survey.
\newblock \emph{AI Open}, 3:\penalty0 71--90, 2022.
\newblock ISSN 2666-6510.
\newblock \doi{https://doi.org/10.1016/j.aiopen.2022.03.001}.
\newblock URL \url{https://www.sciencedirect.com/science/article/pii/S2666651022000080}.

\bibitem[Li(2023)]{Li_2023}
Li, Y.
\newblock A practical survey on zero-shot prompt design for in-context learning.
\newblock In \emph{Proceedings of the Conference Recent Advances in Natural Language Processing - Large Language Models for Natural Language Processings}, RANLP. INCOMA Ltd., Shoumen, BULGARIA, 2023.
\newblock \doi{10.26615/978-954-452-092-2_069}.
\newblock URL \url{http://dx.doi.org/10.26615/978-954-452-092-2_069}.

\bibitem[Li et~al.(2023)Li, Zhu, Lu, and Yin]{li2023synthetic}
Li, Z., Zhu, H., Lu, Z., and Yin, M.
\newblock Synthetic data generation with large language models for text classification: Potential and limitations, 2023.

\bibitem[Liang et~al.(2023)Liang, Bommasani, Lee, Tsipras, Soylu, Yasunaga, Zhang, Narayanan, Wu, Kumar, Newman, Yuan, Yan, Zhang, Cosgrove, Manning, Ré, Acosta-Navas, Hudson, Zelikman, Durmus, Ladhak, Rong, Ren, Yao, Wang, Santhanam, Orr, Zheng, Yuksekgonul, Suzgun, Kim, Guha, Chatterji, Khattab, Henderson, Huang, Chi, Xie, Santurkar, Ganguli, Hashimoto, Icard, Zhang, Chaudhary, Wang, Li, Mai, Zhang, and Koreeda]{liang2023holistic}
Liang, P., Bommasani, R., Lee, T., Tsipras, D., Soylu, D., Yasunaga, M., Zhang, Y., Narayanan, D., Wu, Y., Kumar, A., Newman, B., Yuan, B., Yan, B., Zhang, C., Cosgrove, C., Manning, C.~D., Ré, C., Acosta-Navas, D., Hudson, D.~A., Zelikman, E., Durmus, E., Ladhak, F., Rong, F., Ren, H., Yao, H., Wang, J., Santhanam, K., Orr, L., Zheng, L., Yuksekgonul, M., Suzgun, M., Kim, N., Guha, N., Chatterji, N., Khattab, O., Henderson, P., Huang, Q., Chi, R., Xie, S.~M., Santurkar, S., Ganguli, S., Hashimoto, T., Icard, T., Zhang, T., Chaudhary, V., Wang, W., Li, X., Mai, Y., Zhang, Y., and Koreeda, Y.
\newblock Holistic evaluation of language models, 2023.

\bibitem[Liu et~al.(2024)Liu, Wei, Liu, Si, Zhang, Rao, Zheng, Peng, Yang, Zhou, and Dai]{liu2024best}
Liu, R., Wei, J., Liu, F., Si, C., Zhang, Y., Rao, J., Zheng, S., Peng, D., Yang, D., Zhou, D., and Dai, A.~M.
\newblock Best practices and lessons learned on synthetic data for language models, 2024.

\bibitem[Liu et~al.(2019)Liu, Ott, Goyal, Du, Joshi, Chen, Levy, Lewis, Zettlemoyer, and Stoyanov]{liu2019roberta}
Liu, Y., Ott, M., Goyal, N., Du, J., Joshi, M., Chen, D., Levy, O., Lewis, M., Zettlemoyer, L., and Stoyanov, V.
\newblock Roberta: A robustly optimized bert pretraining approach, 2019.

\bibitem[Mikołajczyk \& Grochowski(2018)Mikołajczyk and Grochowski]{imageaug2}
Mikołajczyk, A. and Grochowski, M.
\newblock Data augmentation for improving deep learning in image classification problem.
\newblock In \emph{2018 International Interdisciplinary PhD Workshop (IIPhDW)}, pp.\  117--122, 2018.
\newblock \doi{10.1109/IIPHDW.2018.8388338}.

\bibitem[Okur et~al.(2022{\natexlab{a}})Okur, Sahay, and Nachman]{okur-etal-2022-data}
Okur, E., Sahay, S., and Nachman, L.
\newblock Data augmentation with paraphrase generation and entity extraction for multimodal dialogue system.
\newblock In Calzolari, N., B{\'e}chet, F., Blache, P., Choukri, K., Cieri, C., Declerck, T., Goggi, S., Isahara, H., Maegaard, B., Mariani, J., Mazo, H., Odijk, J., and Piperidis, S. (eds.), \emph{Proceedings of the Thirteenth Language Resources and Evaluation Conference}, pp.\  4114--4125, Marseille, France, June 2022{\natexlab{a}}. European Language Resources Association.
\newblock URL \url{https://aclanthology.org/2022.lrec-1.437}.

\bibitem[Okur et~al.(2022{\natexlab{b}})Okur, Sahay, and Nachman]{okur2022data}
Okur, E., Sahay, S., and Nachman, L.
\newblock Data augmentation with paraphrase generation and entity extraction for multimodal dialogue system, 2022{\natexlab{b}}.

\bibitem[Reimers \& Gurevych(2019)Reimers and Gurevych]{reimers2019sentencebert}
Reimers, N. and Gurevych, I.
\newblock Sentence-bert: Sentence embeddings using siamese bert-networks, 2019.

\bibitem[Saravia et~al.(2018)Saravia, Liu, Huang, Wu, and Chen]{emo}
Saravia, E., Liu, H.-C.~T., Huang, Y.-H., Wu, J., and Chen, Y.-S.
\newblock {CARER}: Contextualized affect representations for emotion recognition.
\newblock In \emph{Proceedings of the 2018 Conference on Empirical Methods in Natural Language Processing}, pp.\  3687--3697, Brussels, Belgium, October-November 2018. Association for Computational Linguistics.
\newblock \doi{10.18653/v1/D18-1404}.
\newblock URL \url{https://www.aclweb.org/anthology/D18-1404}.

\bibitem[Socher et~al.(2013)Socher, Perelygin, Wu, Chuang, Manning, Ng, and Potts]{sst2}
Socher, R., Perelygin, A., Wu, J., Chuang, J., Manning, C.~D., Ng, A., and Potts, C.
\newblock Recursive deep models for semantic compositionality over a sentiment treebank.
\newblock In Yarowsky, D., Baldwin, T., Korhonen, A., Livescu, K., and Bethard, S. (eds.), \emph{Proceedings of the 2013 Conference on Empirical Methods in Natural Language Processing}, pp.\  1631--1642, Seattle, Washington, USA, October 2013. Association for Computational Linguistics.
\newblock URL \url{https://aclanthology.org/D13-1170}.

\bibitem[Song et~al.(2022)Song, Wang, Mondal, and Sahoo]{song2022comprehensive}
Song, Y., Wang, T., Mondal, S.~K., and Sahoo, J.~P.
\newblock A comprehensive survey of few-shot learning: Evolution, applications, challenges, and opportunities, 2022.

\bibitem[Steck et~al.(2024)Steck, Ekanadham, and Kallus]{Steck2024CosineSimilarity}
Steck, H., Ekanadham, C., and Kallus, N.
\newblock Is cosine-similarity of embeddings really about similarity?
\newblock \emph{arXiv preprint arXiv:2403.05440v1}, 2024.
\newblock arXiv.org perpetual non-exclusive license.

\bibitem[Stefano(2021)]{nyt}
Stefano, D.~D.
\newblock New york times topics.
\newblock \url{https://huggingface.co/datasets/dstefa/New_York_Times_Topics}, 2021.

\bibitem[Veselovsky et~al.(2023)Veselovsky, Ribeiro, Arora, Josifoski, Anderson, and West]{veselovsky2023generating}
Veselovsky, V., Ribeiro, M.~H., Arora, A., Josifoski, M., Anderson, A., and West, R.
\newblock Generating faithful synthetic data with large language models: A case study in computational social science, 2023.

\bibitem[Wang et~al.(2019)Wang, Pruksachatkun, Nangia, Singh, Michael, Hill, Levy, and Bowman]{wang2019superglue}
Wang, A., Pruksachatkun, Y., Nangia, N., Singh, A., Michael, J., Hill, F., Levy, O., and Bowman, S.~R.
\newblock Super{GLUE}: A stickier benchmark for general-purpose language understanding systems.
\newblock \emph{arXiv preprint 1905.00537}, 2019.

\bibitem[Wang et~al.(2023)Wang, Zhou, and Sachan]{wang2023lets}
Wang, R., Zhou, W., and Sachan, M.
\newblock Let's synthesize step by step: Iterative dataset synthesis with large language models by extrapolating errors from small models, 2023.

\bibitem[Wang et~al.(2020)Wang, Zhang, and Wu]{wang2020structuralaware}
Wang, Z., Zhang, Y., and Wu, H.
\newblock Structural-aware sentence similarity with recursive optimal transport, 2020.

\bibitem[Wei et~al.(2023)Wei, Wang, Schuurmans, Bosma, Ichter, Xia, Chi, Le, and Zhou]{COT}
Wei, J., Wang, X., Schuurmans, D., Bosma, M., Ichter, B., Xia, F., Chi, E., Le, Q., and Zhou, D.
\newblock Chain-of-thought prompting elicits reasoning in large language models, 2023.

\bibitem[Xie et~al.(2020)Xie, Dai, Hovy, Luong, and Le]{xie2020unsupervised}
Xie, Q., Dai, Z., Hovy, E., Luong, M.-T., and Le, Q.~V.
\newblock Unsupervised data augmentation for consistency training, 2020.

\bibitem[Yang et~al.(2023)Yang, Xiao, Zhang, Guo, Zhao, and Shen]{cvaug}
Yang, S., Xiao, W., Zhang, M., Guo, S., Zhao, J., and Shen, F.
\newblock Image data augmentation for deep learning: A survey, 2023.

\bibitem[Yang et~al.(2020)Yang, Malaviya, Fernandez, Swayamdipta, Le~Bras, Wang, Bhagavatula, Choi, and Downey]{nlpaug2}
Yang, Y., Malaviya, C., Fernandez, J., Swayamdipta, S., Le~Bras, R., Wang, J.-P., Bhagavatula, C., Choi, Y., and Downey, D.
\newblock Generative data augmentation for commonsense reasoning.
\newblock In Cohn, T., He, Y., and Liu, Y. (eds.), \emph{Findings of the Association for Computational Linguistics: EMNLP 2020}, pp.\  1008--1025, Online, November 2020. Association for Computational Linguistics.
\newblock \doi{10.18653/v1/2020.findings-emnlp.90}.
\newblock URL \url{https://aclanthology.org/2020.findings-emnlp.90}.

\bibitem[Ye et~al.(2022)Ye, Gao, Li, Xu, Feng, Wu, Yu, and Kong]{ye2022zerogen}
Ye, J., Gao, J., Li, Q., Xu, H., Feng, J., Wu, Z., Yu, T., and Kong, L.
\newblock Zerogen: Efficient zero-shot learning via dataset generation, 2022.

\bibitem[Yu et~al.(2023)Yu, Zhuang, Zhang, Meng, Ratner, Krishna, Shen, and Zhang]{attrgen}
Yu, Y., Zhuang, Y., Zhang, J., Meng, Y., Ratner, A., Krishna, R., Shen, J., and Zhang, C.
\newblock Large language model as attributed training data generator: A tale of diversity and bias, 2023.

\bibitem[Zhou et~al.(2024)Zhou, Guo, Wang, Chang, and Wu]{zhou2024survey}
Zhou, Y., Guo, C., Wang, X., Chang, Y., and Wu, Y.
\newblock A survey on data augmentation in large model era, 2024.

\end{thebibliography}
\bibliographystyle{icml2024}

\newpage
\appendix
\onecolumn
\section{Appendix}
\label{appendix:a}

Prompt for topic generation for zero-shot with topics and LLM output examples. GPT-4 is used to generate 500 random topics per task:
\begin{table}[h]  
\begin{center}
\begin{tabular}{llp{10cm}}    
\toprule
Task & Role & Message \\   
\midrule
BoolQ, RET, NYT, SST-2, Emo & System & You are an AI assistant that generates random topics. There is no limit on the number of topics you can generate. \\ 
BoolQ, RET, NYT & User & Please generate 500 topics \\    
BoolQ, RET, NYT & LLM & Output example: The world's most beautiful sculptures, The role of technology in modern education ... \\   
SST-2, Emo & User & Please generate 500 twitter post topics \\    
SST-2, Emo & LLM & Output example: Lunch break, Online dating ... \\    
Review & System & You are an AI assistant that knows Amazon product categories. The user will ask you to generate a list of categories. It is your responsibility to generate the entire list of categories. \\     
Review & User & Please generate 500 amazon different product categories \\     
Review & LLM & Output example: Baby Products, Clothing, Jewelry ... \\  
\bottomrule
\end{tabular}    
\label{table: topicgenexample}
\end{center}
\end{table}    

\section{Appendix}
\label{appendix:b}
Prompt for Question Rephrasing in Section 5.2  
\begin{table}[h]    
\vskip 0.15in    
\begin{center}    
\begin{small}    
\begin{tabular}{p{\linewidth}}    
\toprule  
Please rephrase the question as if you are typing it in a search engine. Make sure the answer can only be true or false, Input: {question} Output: \\  
\bottomrule  
\end{tabular}    
\end{small}    
\end{center}    
\vskip -0.1in    
\end{table}  

\clearpage
\section{Appendix}
\label{appendix:c}

Prompt used for data generation for each task:
\begin{table}[h]    
\begin{tabular}{lcp{10cm}}    
\toprule
\textbf{Task} & \textbf{Prompt Type} & \textbf{Prompt} \\   
\midrule
BoolQ & zero-shot & \paragraph{Step 1} Please generate a random short passage. Passage: \paragraph{Step 2} Please generate a True or False question based on the passage. The answer to the question must be [random([True, False])] Passage: [passage from step 1] Question: \\ 
BoolQ & zero-shot topic & \paragraph{Step 1} Please generate a short passage about this topic: [topic sampled from a topic list] Passage: \paragraph{Step 2} Please generate a True or False question based on the passage. The answer to the question must be [random([True, False])] Passage: [passage from step 1] Question: \\ 
BoolQ & one-shot & \paragraph{Step 1} Please generate a Passage, a Question and the Label to the question following this example: [example from raw data: Passage, Question, Label] Please generate a similar passage. Passage: \paragraph{Step 2} Please generate a True or False question based on the passage. The answer to the question must be [label from example in Step 1] Passage: [passage generated in Step 1] Question:\\ 
BoolQ & few-shot (3 or 5) & \paragraph{Step 1} Please generate a Passage, a Question and the Label to the question. Here are some examples: [examples from raw data: Passage, Question, Label] Please generate a similar example. Make sure the question is a True or False question and the answer to the question is [random([True, False])]. Passage: \\
EMO & zero-shot & \paragraph{Step 1} Please generate a twitter post with the emotion of [random(label)]. Text: \\  
EMO & zero-shot topic & \paragraph{Step 1} Please consider this topic for generation: [topic sampled from a topic list]. Please generate a twitter post with the emotion of [random(label)]. Text: \\   
EMO & one-shot & \paragraph{Step 1} The task is to predict the emotion of a twitter post. The emotion contains six categories: sadness, joy, love, anger, fear, surprise. Here is an example. Text: [example from raw data] Emotion: [example label from raw data] Please generate another example for the same emotion. Text: \\  
EMO & few-shot (3 or 5)& \paragraph{Step 1} The task is to predict the emotion of a twitter post. The emotion contains six categories: sadness, joy, love, anger, fear, surprise. Here are some examples: [examples: Text, Emotion] Please generate a twitter post with the emotion of [first label from examples]. Text: \\ 
\bottomrule
\end{tabular}    
\label{tab:messages_tasks}    
\end{table}  

\begin{table}    
\begin{tabular}{lcp{10cm}}  
\toprule
Task & Prompt Type &Prompt \\ 
\midrule
NYT & zero-shot & \paragraph{Step 1} Please generate a news title for [random(label)] category. Headline: \\  
NYT & zero-shot topic & \paragraph{Step 1} Please consider this sentence for generation: [topic sampled from topic list]. Please generate a news headline for [random(label)] category. Headline: \\ 
NYT & one-shot & \paragraph{Step 1} The task is to predict the topic of a news headline. The topics contain 'sports', 'arts, culture and entertainment', 'business and finance', 'health and wellness', 'lifestyle and fashion', 'science and technology', 'politics', 'crime'. Here is an example News: [example news] Topic: [example topic] Please generate another news on [example topic]. Headline: \\ 
NYT & few-shot (3 or 5) & \paragraph{Step 1} The task is to predict the topic of a news headline. The topics contain 'sports', 'arts, culture and entertainment', 'business and finance', 'health and wellness', 'lifestyle and fashion', 'science and technology', 'politics', 'crime'. Here are some examples: [examples: Headline, Topic] Please generate a news headline for [first topic from examples] category. News: \\ 
Review & zero-shot & \paragraph{Step 1} The Amazon customer review has a rating ranges from 1 to 5, 1 being the lowest and 5 being the highest. Please generate a customer review with a rating of [random(label)]. Content: \\ 
Review & zero-shot topic & \paragraph{Step 1} The Amazon customer review has a rating ranges from 1 to 5, 1 being the lowest and 5 being the highest. Please generate a customer review with a rating of [random(label)] for a specific product under [a product category sampled from topic list]. Please use a fake product name. Content: \\ 
Review & one-shot & \paragraph{Step 1} The task is to predict the rating of an Amazon customer review based on the content. The rating ranges from 1 to 5, 1 being the lowest and 5 being the highest. Here is a review example. Content: [example content] Rating: [example rating] Please generate another example for a similar product. Make sure the rating for the review is [example rating]. Content: \\ 
Review & few-shot (3 or 5) & \paragraph{Step 1} The Amazon customer review has a rating ranges from 1 to 5, 1 being the lowest and 5 being the highest. Here are some examples Content: [examples: Content, Rating] Please generate a customer review with a rating [first rating from examples]. Content: \\ 
\bottomrule
\end{tabular}    
\label{tab:messages_tasks}    
\end{table}    

\begin{table}    
\begin{tabular}{lcp{10cm}}    
\toprule
Task&Prompt Type &Prompt \\ 
\midrule
RTE & zero-shot & \paragraph{Step 1} Given a premise and a hypothesis, a model needs to predict whether the hypothesis can be logically inferred from the premise. The response should be either True if the hypothesis can be inferred from the premise, or False if it cannot be inferred. Here is the output format: Premise: Hypothesis: Label: True or False Please generate an example where the Label is [random(label)]. Premise: \\ 
RTE & zero-shot topic & \paragraph{Step 1} Given a premise and a hypothesis, a model needs to predict whether the hypothesis can be logically inferred from the premise. The response should be either True if the hypothesis can be inferred from the premise, or False if it cannot be inferred. Here is the output format: Premise: Hypothesis: Label: True or False Please generate an example about [premise] where the Label is [random(label)]. Premise: \\ 
RTE & one-shot & \paragraph{Step 1} Given a premise and a hypothesis, a model needs to predict whether the hypothesis can be logically inferred from the premise. The response should be either True if the hypothesis can be inferred from the premise, or False if it cannot be inferred. Here is an example: Premise: [example premise] Hypothesis: [example hypothesis] Label: [example label] Please generate another similar example where the Label is [example label]. Premise: \\ 
RTE & few-shot  (3 or 5)& \paragraph{Step 1} Given a premise and a hypothesis, a model needs to predict whether the hypothesis can be logically inferred from the premise. The response should be either True if the hypothesis can be inferred from the premise, or False if it cannot be inferred. Here are some examples: [examples: Premise, Hypothesis, Label] Please generate a similar example. Make sure the label is [first label from examples]. Premise: \\   

SST-2 & zero-shot & \paragraph{Step 1} Please generate a sentence that contains a [random(label)] sentiment. Sentence: \\   
SST-2 & zero-shot topic & \paragraph{Step 1} Please consider this topic for generation: [topic from the topic list]. Please generate a sentence that contains a [random(label)]
sentiment. Sentence: \\ 
SST-2 & one-shot & \paragraph{Step 1} The task is to predict whether the following sentence is positive or negative sentiment. Sentence: [example sentence] Label:[example label] Please generate a similar example on the same topic, including a Sentence and a Label. Sentence: \\  
SST-2 & few-shot  (3 or 5)& \paragraph{Step 1} The task is to predict whether the following sentence is positive or negative sentiment. [examples: Sentence, Label] Please generate a similar example, including a Sentence and a Label. Sentence: \\ 
\bottomrule
\end{tabular}    
\label{tab:messages_tasks}    
\end{table}
\clearpage
\section{Appendix}
\label{appendix:d}
Prompt used to evaluate LLM performance on each task.
\begin{table}[h]    
\begin{tabular}{lcp{10cm}}   
\toprule
Task& Prompt Type &Prompt \\ 
\midrule
RTE & zero-shot & \paragraph{Step 1} Given a premise and a hypothesis, a model needs to predict whether the hypothesis can be logically inferred from the premise. The response should be either True if the hypothesis can be inferred from the premise, or False if it cannot be inferred. Premise: [premise], Hypothesis: [hypothesis], Label:\\  
RTE & 0/1/3/5-shot & \paragraph{Step 1} Given a premise and a hypothesis, a model needs to predict whether the hypothesis can be logically inferred from the premise. The response should be either True if the hypothesis can be inferred from the premise, or False if it cannot be inferred. Here are some examples: [example premise, hypothesis, label] Premise: [premise], Hypothesis: [hypothesis], Label: \\    

BoolQ & zero-shot & \paragraph{Step 1} The task is to answer a question which is solely based on the content provided. Passage: [passage] , Question: [question], Label: \\  

BoolQ & 0/1/3/5-shot & \paragraph{Step 1} The task is to answer a question which is solely based on the content provided. Here are some examples: [example passage, question, label] Passage: [passage], Question: [question], Label: \\  

Review & zero-shot & \paragraph{Step 1} The task is to predict the rating of an Amazon customer review based on the content. The rating ranges from 1 to 5, with 1 being the lowest and 5 being the highest. Text: [text] , Label: \\  

Review & 0/1/3/5-shot & \paragraph{Step 1} The task is to predict the rating of an Amazon customer review based on the content. The rating ranges from 1 to 5, with 1 being the lowest and 5 being the highest. Here are some examples: [example text, label] Text: [text], Label: \\  

NYT & zero-shot & \paragraph{Step 1} The task is to predict the topic of a news headline. The topics include: 'sports', 'arts, culture and entertainment', 'business and finance', 'health and wellness', 'lifestyle and fashion', 'science and technology', 'politics', 'crime'. Text:[text], Label: \\  

NYT & 0/1/3/5-shot & \paragraph{Step 1} The task is to predict the topic of a news headline. The topics include: 'sports', 'arts, culture and entertainment', 'business and finance', 'health and wellness', 'lifestyle and fashion', 'science and technology', 'politics', 'crime'. Here are some examples: [example text, label] Text: [text], Label: \\  

EMO & zero-shot & \paragraph{Step 1} The task is to predict the emotion of a Twitter text. The emotions include six categories: sadness, joy, love, anger, fear, surprise. Text: [text], Label: \\  

EMO & 0/1/3/5-shot & \paragraph{Step 1} The task is to predict the emotion of a Twitter text. The emotions include six categories: sadness, joy, love, anger, fear, surprise. Here are some examples: [example text, label] Text: [text], Label:\\  
SST-2 & zero-shot & \paragraph{Step 1} The task is to predict whether the given sentence has a positive or negative sentiment. Sentence: [sentence], Label:\\ 

SST-2 & 0/1/3/5-shot & \paragraph{Step 1} The task is to predict whether the given sentence has a positive or negative sentiment. Here are some examples: [example sentence, label], Sentence: [sentence], Label:\\ 
\bottomrule
\end{tabular}    
\label{tab:llm_task_prompt}    
\end{table}
\end{document}